\documentclass[]{jingdong}
\usepackage[toc,page,header]{appendix}
\usepackage{latexsym}
\usepackage[T1]{fontenc}
\usepackage[utf8]{inputenc}
\usepackage{microtype}
\usepackage{inconsolata}
\usepackage{graphicx}
\usepackage{hyperref}       
\usepackage{url}            
\usepackage{booktabs}       
\usepackage{amsfonts}       
\usepackage{nicefrac}       
\usepackage{stackengine}
\usepackage{microtype}      
\usepackage{colortbl}
\usepackage{xcolor}
\usepackage{amsmath}
\usepackage{amssymb}
\usepackage{amsthm}
\usepackage{mathrsfs}
\usepackage{pifont}
\usepackage{MnSymbol}
\usepackage{balance}
\usepackage{enumitem}
\usepackage{listings}
\usepackage{xcolor}
\usepackage{natbib}
\usepackage{multicol}

\AtBeginDocument{%
  \providecommand\BibTeX{{%
    \normalfont B\kern-0.5em{\scshape i\kern-0.25em b}\kern-0.8em\TeX}}}

\makeatletter
\DeclareRobustCommand\onedot{\futurelet\@let@token\@onedot}
\def\@onedot{\ifx\@let@token.\else.\null\fi}

\usepackage{setspace}
\usepackage{mathtools}

\usepackage{multirow,booktabs}
\usepackage{subcaption}

\newcommand{\owo}[1]{\textsc{OAgents}}

\definecolor{lightgreen}{RGB}{144, 238, 144} 
\definecolor{lightred}{RGB}{255, 105, 97}

\newtcolorbox{promptbox}[2][Prompt]{
colback=black!5!white,
arc=5pt, 
boxrule=0.5pt,
fonttitle=\bfseries,
title=#1, 
before upper={\small}, fontupper=\fontfamily{ptm}\selectfont,
colframe=#2, 
}
\definecolor{ogreen}{RGB}{34, 139, 34}

\usepackage{cleveref}
\theoremstyle{plain}

\theoremstyle{definition}

\theoremstyle{remark}
\usepackage{subcaption}

\usepackage[textsize=tiny]{todonotes}
\usepackage{fontawesome}

\usepackage{xcolor}
\usepackage{graphicx}

\usepackage{wrapfig}
\usepackage[edges]{forest}

\usepackage{amsmath}

\usepackage{makecell} 
\usepackage{minitoc}
\usepackage{multirow} 
\usepackage[most]{tcolorbox}
\usepackage{enumitem}
\usepackage{listings}         
\tcbuselibrary{listings} 
\usepackage{amssymb}
\usepackage{wrapfig}
\usepackage{forest}

\useforestlibrary{edges}
\usepackage{ifthen}

\title{Harnessing Streaming Video in the Wild}
\author[1,2,*]{Dingyu Yao}
\author[1,2,*]{Shuhuan Gu}
\author[3,*]{Qingyi Si}
\author[1,2,*]{Junhao Zhou}
\author[1,2]{\ \ \ \ \ \ \ \ \ \ \ \ \ \ \ Chenxu Yang}
\author[1,2]{Chuanyu Qin}
\author[1,2]{Naibin Gu}
\author[1,2 , \dagger]{Zheng Lin}
\author[1]{Weiping Wang}
\author[3]{Nan Duan}
\author[3]{Jiaqi Wang}

\affiliation[1]{Institute of Information Engineering, Chinese Academy of Sciences, Beijing, China}
\affiliation[2]{School of Cyber Security, University of Chinese Academy of Sciences, Beijing, China}
\affiliation[3]{JD.COM}

\contribution[*]{Equal contribution.}
\contribution[\dagger]{Corresponding author.}

\abstract{
Vision-Language Models (VLMs) are increasingly required to process unbounded video streams in applications such as video-call assistants, live commentary, and embodied robots. An ideal streaming system should support proactive interaction, long-horizon memory, and real-time processing, while resting on a VLM backbone capable of handling diverse in-the-wild streaming tasks. However, existing VLMs excel at offline video understanding but fall short in streaming capabilities and lack dedicated infrastructure for streaming deployment. We address this gap on three fronts. (i) For backbone capability, we construct \textbf{Streaming-Train-248K}, a streaming dataset paired with a novel training objective for adapting VLMs to streaming interaction and understanding. (ii) For real-world deployment, we introduce \textbf{Streaming Harness}, a plug-and-play system that endows any VLM with three core abilities: proactive interaction (per-second response decisions), long-term memory (12-hour context retention), and real-time processing (sub-second latency). (iii) To drive community progress on real-streaming capabilities, we design \textbf{Streaming-Eval}, a benchmark that reflects models' capabilities across diverse in-the-wild scenarios. Extensive experiments demonstrate consistent gains across all four streaming capabilities. We will open-source our data, harness, and benchmark to accelerate the community's transition from offline video understanding to deployable streaming intelligence. 
}

\checkdata[Email]{\email{yaodingyu@iie.ac.cn}; \email{linzheng@iie.ac.cn}; \email{siqingyi.phoebus@jd.com}}
\begin{document}
\maketitle

\section{Introduction}
\begin{figure}[!h]
  \centering
  \includegraphics[width=1\textwidth]{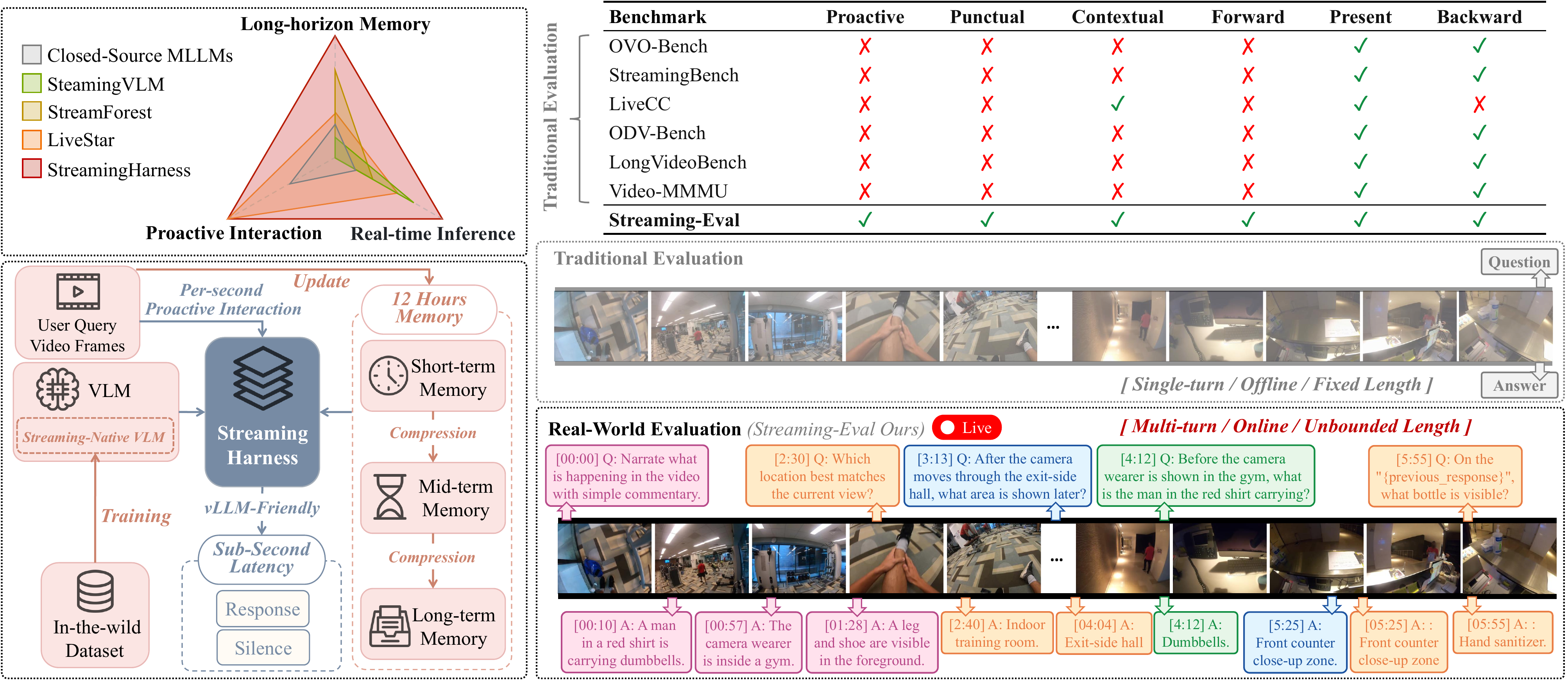}
  \caption{\textbf{Top-left:}  StreamingHarness jointly delivers \textit{proactive interaction}, \textit{long-horizon memory}, and \textit{real-time inference}—capabilities that no prior baseline satisfies simultaneously. \textbf{Bottom-left:} Overview of the StreamingHarness. \textbf{ Top-right:} Comparative overview of Streaming-Eval and existing benchmarks. \textbf{Bottom-right:} Unlike single-turn offline benchmarks, Streaming-Eval assesses six in-the-wild capabilities (\textit{Proactive}, \textit{Punctual}, \textit{Contextual}, \textit{Forward}, \textit{Present}, \textit{Backward}).}
  \label{fig:intro}
\end{figure}
Vision-Language Models (VLMs) ~\cite{bai2025qwen3vltechnicalreport,gemini3pro2025,kimiteam2026kimik25visualagentic,seed2_2026} have substantially advanced video understanding, achieving strong results on offline tasks~\cite{fu2026videommev2stagebenchmarkscomprehensive,wang2025lvbenchextremelongvideo,10815993,10678476}, such as video question answering~\cite{fu2026videommev2stagebenchmarkscomprehensive,wang2025lvbenchextremelongvideo,10815993}, temporal grounding~\cite{10678476}, and long-form captioning~\cite{10815993}. 
These tasks share a common formulation: the model jointly attends over a complete, pre-recorded clip and produces a one-shot response. 
However, a growing class of real-world applications fundamentally violates this assumption. 
Yet many real-world applications violate this assumption.
In video-call assistants~\cite{seed2_2026}, live commentary~\cite{11092811,10.1109/TCSVT.2024.3462433},
online surveillance~\cite{10.1109/TCSVT.2024.3462433},
and embodied robots~\cite{li2026egolivelargescaleegocentricdataset}, video arrives as an unbounded stream, and the model is required to respond causally, conditioning solely on previously observed frames under stringent latency constraints. 

An ideal streaming system should satisfy four fundamental requirements: \textit{(i) a versatile VLM backbone} capable of handling diverse in-the-wild streaming tasks; \textit{(ii) proactive interaction}, whereby the model determines when to respond and when to remain silent, generating outputs causally without access to future frames; \textit{(iii) long-horizon memory}, which enables the retrieval of historical evidence frames pertinent to a given query across streams of unbounded length; and \textit{(iv) real-time inference}, ensuring that responses are produced under low-latency constraints and remain temporally synchronized with the incoming input.
A growing body of work has sought to advance streaming video understanding along these axes~\cite{wang-etal-2025-videollm,wang2026streambridge,zeng2026streamforest,yang2025livestar,10657274,11094534,zhang2025flashvstreamefficientrealtimeunderstanding,zhang2026hermeskvcachehierarchical,Ding_2025_ICCV,xu2025streamingvlmrealtimeunderstandinginfinite}; however, as illustrated in Figure~\ref{fig:intro} (top-left), most existing methods address only one or two of them in isolation, rather than jointly satisfying all of them within a unified framework.
To close this gap, we present a unified stack across three fronts (model, harness and benchmark) toward an ideal streaming system.

For backbone capability, we train a Streaming-Native VLM for proactive interaction, capable of handling diverse 
\textbf{\textit{in-the-wild streaming tasks}}. 
We introduce two special tokens~\cite{11094534,Ding_2025_ICCV}, \texttt{</response>} and \texttt{</silence>}, which enable the model to decide whether to respond based on the content observed so far.
To support this, we carefully construct \textbf{Streaming-Train-248K}, a high-quality dataset spanning a rich variety of real-world task types. 
The dataset provides per-second frame-text alignment, pairing each one-second segment with either an explicit response or a silence token.
Together with a tailored loss function that mitigates the silence-response imbalance, we achieve more stable training.

For real-world deployment, we propose \textbf{StreamingHarness}—a plug-and-play system that instantly equips any VLM with strong instruction-following capabilities for streaming video processing.
Concretely, as illustrated in Figure~\ref{fig:intro} (bottom-left), the harness manages \textbf{\textit{long-horizon memory}} through a three-tier memory management module, supporting up to 12 hours of context.
In parallel, a prefix-aware Key-Value (KV) cache design makes the harness compatible with native \texttt{vLLM}, allowing KV cache reuse to preserve sub-second latency and thus deliver \textbf{\textit{real-time processing}}. 
Furthermore, to support \textbf{\textit{proactive interaction}}, user queries may be issued at any moment, and the harness invokes the backbone every second, letting the model decide whether to respond at each tick. The dedicated system prompt can be extended to any VLM that is not natively trained for streaming tasks.

Model capability and deployment infrastructure are necessary, yet without a principled way to measure whether they jointly meet genuine streaming requirements, progress is hard to track. 
As shown in Figure~\ref{fig:intro} (top-right), existing benchmarks~\cite{zeng2026streamforest,11463959,11094230,wang2025omnimmicomprehensivemultimodalinteraction,11093281,wu2024longvideobenchbenchmarklongcontextinterleaved,fu2025videommefirstevercomprehensiveevaluation,hu2025videommmuevaluatingknowledgeacquisition} reduce evaluation to single-turn, offline-style queries over short clips, overlooking core streaming behaviors, such as recalling distant evidence, deferring responses pending future evidence, and maintaining coherence across the stream. 
To this end, we design \textbf{Streaming-Eval}, a benchmark spanning two in-the-wild axes: Streaming Interaction, which evaluates whether (Proactive), when (Punctual), and how coherently across interdependent turns (Contextual) a model responds; and Streaming Understanding, which evaluates reasoning over past (Backward), ongoing (Present), and future (Forward) visual evidence.
We further propose SW-F1 (Streaming Weighted F1), a metric that jointly captures answer correctness and response timeliness.

We evaluate our stack on Streaming-Eval, spanning narration and question answering.
Our training (Section~\ref{sec:4}) transforms an offline VLM into a streaming-native model, yielding substantial gains in proactive interaction.
Equipping off-the-shelf VLMs with StreamingHarness (Section~\ref{sec:5}) further endows them with long-horizon memory, particularly improving performance on backward tasks.
Combining our streaming-native 8B VLM with StreamingHarness achieves state-of-the-art results, surpassing most strong closed-source models.
Moreover, thanks to our vLLM-friendly design, our system sustains coherent narration over two-hour live sports broadcasts at stable sub-second latency. Below are the key contributions of our work:
\setlength{\itemsep}{0pt}
\begin{itemize}
    \item \textbf{Streaming-Train-248K}. We construct a large-scale streaming dataset with per-second frame-text alignment, paired with a tailored training objective, to adapt VLMs to \textit{in-the-wild} streaming interaction and yield streaming-native VLMs.
    \item  \textbf{StreamingHarness}. We introduce a plug-and-play framework that equips any VLM with three core streaming abilities: \textit{proactive interaction} (per-second response decisions), \textit{long-horizon memory} (12-hour context), and \textit{real-time inference} (sub-second latency).
    \item \textbf{Streaming-Eval}. We design a benchmark targeting in-the-wild streaming tasks, which captures the capabilities required for real-world deployment, encouraging the community to shift from offline video understanding toward deployable streaming intelligence.
    \item \textbf{Excellent Performance}. Experiments show that our 8B streaming-native model equipped with StreamingHarness achieves state-of-the-art results, significantly outperforming Claude Opus 4.6, GPT 5.4, Gemini 3.1 Pro, and Doubao Seed 2.0 Pro. 
\end{itemize}
\section{Related Works}
\label{related}

\subsection{Streaming Video Understanding}
Beyond a versatile VLM backbone, streaming video understanding also demands proactive interaction, long-horizon memory, and real-time inference. However, most existing methods address only one or two of these capabilities in isolation: 
\textbf{\textit{Proactive Interaction.}}
A core challenge in streaming video understanding is deciding whether and when to respond under continuous input.
VideoLLM-online~\cite{10657274} pioneers training the model to emit EOS on redundant frames and respond on informative ones.
Later works explore alternative designs, including a decision token~\cite{yan2026proactvl}, response-state tokens unified with content generation~\cite{xia2025streamingvideoinstructiontuning}, perplexity-based gating~\cite{yang2025livestar}, and asynchronous decoupling of perception, decision, and reaction~\cite{qian2025dispider}.
\textbf{\textit{Long-horizon Memory.}}
As streaming videos are unbounded in length, retaining all visual tokens incurs prohibitive memory and compute costs, motivating memory management. Some methods optimize visual tokens or the KV cache, e.g., redundant token pruning~\cite{yao2025timechatonline,wang2026acceleratingstreamingvideolarge}, KV cache offloading~\cite{di2025streaming,yao-etal-2025-tailorkv,11408603}, and hierarchical KV cache memory~\cite{zhang2026hermeskvcachehierarchical,yao2025vecinferefficientllminference}; others~\cite{xiong2025streaming,zhang2025flashvstreamefficientrealtimeunderstanding} introduce dedicated memory architectures that store compressed representations and retrieve relevant content at query time.
\textbf{\textit{Real-time Inference.}}
Unlike offline video understanding, streaming video requires low-latency responses synchronized with incoming frames, necessitating dedicated inference-efficiency optimizations.
Existing efforts focus on the following directions: constraining the context via a fixed-size sliding window~\cite{xu2026streamingvlm,shen2026simplebaselinestreamingvideo}, pruning redundant visual tokens selectively~\cite{xie2026fluxmemadaptivehierarchicalmemory,wu2024videollmmod}, and decoupling perception from response generation through a two-stage pipeline~\cite{Ding_2025_ICCV,li2025lion}.
However, these efficiency gains come at the cost of long-term memory, and current methods struggle to jointly achieve real-time inference and hours-long memory.
Moreover, most existing methods rely on native \texttt{Transformers}~\cite{wolf2020huggingfacestransformersstateoftheartnatural} and are incompatible with efficient inference engines such as \texttt{vLLM}~\cite{kwon2023efficientmemorymanagementlarge} and \texttt{SGLang}~\cite{zheng2024sglangefficientexecutionstructured}, which limits their deployment in real-world production environments.

\subsection{Streaming Video Benchmarks}
As for evaluation, existing mainstream benchmarks~\cite{11092811,11094230,11093281,11463959} primarily assess the correctness of responses. More critically, constrained by online inference frameworks, these benchmarks reduce inference to a single-turn offline-style paradigm, which requires the model to respond immediately, with questions restricted to past or present observed content~\cite{chatterjee2026dontpausepredictionmatters}.
Therefore, a well-designed benchmark is essential for advancing research on streaming video. 
To align with in-the-wild streaming video scenarios, Streaming-Eval evaluates streaming capabilities under online deployment conditions.

\section{Problem Formulation: Streaming as Multi-turn Online Interaction}
\label{sec:3}
We formulate the streaming setting as follows. A video is revealed sequentially as a sequence of temporal segments $\mathcal{V}=\{v_1, v_2, \ldots, v_T\}$, where each segment $v_t$ spans a brief fixed-length interval (e.g., one second) and is anchored by a timestamp token $\tau_t$ of the form \texttt{<$t$s-$(t{+}1)$s>}. The user may issue textual queries at sparsely distributed moments along the timeline; we denote by $q_t$ the query received at step $t$, with $q_t=\varnothing$ when no query is present. At each step, the model emits $a_t$ as either (i) an explicit utterance \texttt{</response>\,answer} when accumulated evidence suffices to address a pending query or narrate a salient event, or (ii) a silence token \texttt{</silence>} otherwise. This formulation promotes temporally grounded responses, suppresses premature predictions, and mitigates hallucinated outputs.
Each step thus constitutes an interaction unit $u_t = (\tau_t,\, v_t,\, q_t,\, a_t)$, and the dialogue history is $\mathcal{H}_{t-1} = \bigl(u_1,\, u_2,\, \ldots,\, u_{t-1}\bigr)$. Conditioned on this history, the response at step $t$ is generated as
$a_t = f_\theta\!\left(\mathcal{H}_{t-1},\, \tau_t,\, v_t,\, q_t\right).$
In contrast to offline video understanding that consumes the entire video a priori, streaming video understanding has no access to future segments $\{v_{t+1}, \ldots, v_T\}$ and must respond causally based on previously observed frames. Through this multi-turn online interaction, the model determines whether to speak at each step $t$.
\section{Streaming-Native VLM in the wild}
\label{sec:4}
As analyzed in Section~\ref{sec:3}, streaming video inference is characterized by multi-turn interaction, where the model decides at each step whether to respond or remain silent. To align with this inference-time schedule, we train a Streaming-Native VLM for proactive interaction by interleaving vision and text tokens during training, and curate a 248K dataset covering diverse proactive perception scenarios.

\subsection{Data Statistics} 
\begin{wrapfigure}{r}{0.5\textwidth}  
    \centering
    \includegraphics[width=0.5\textwidth]{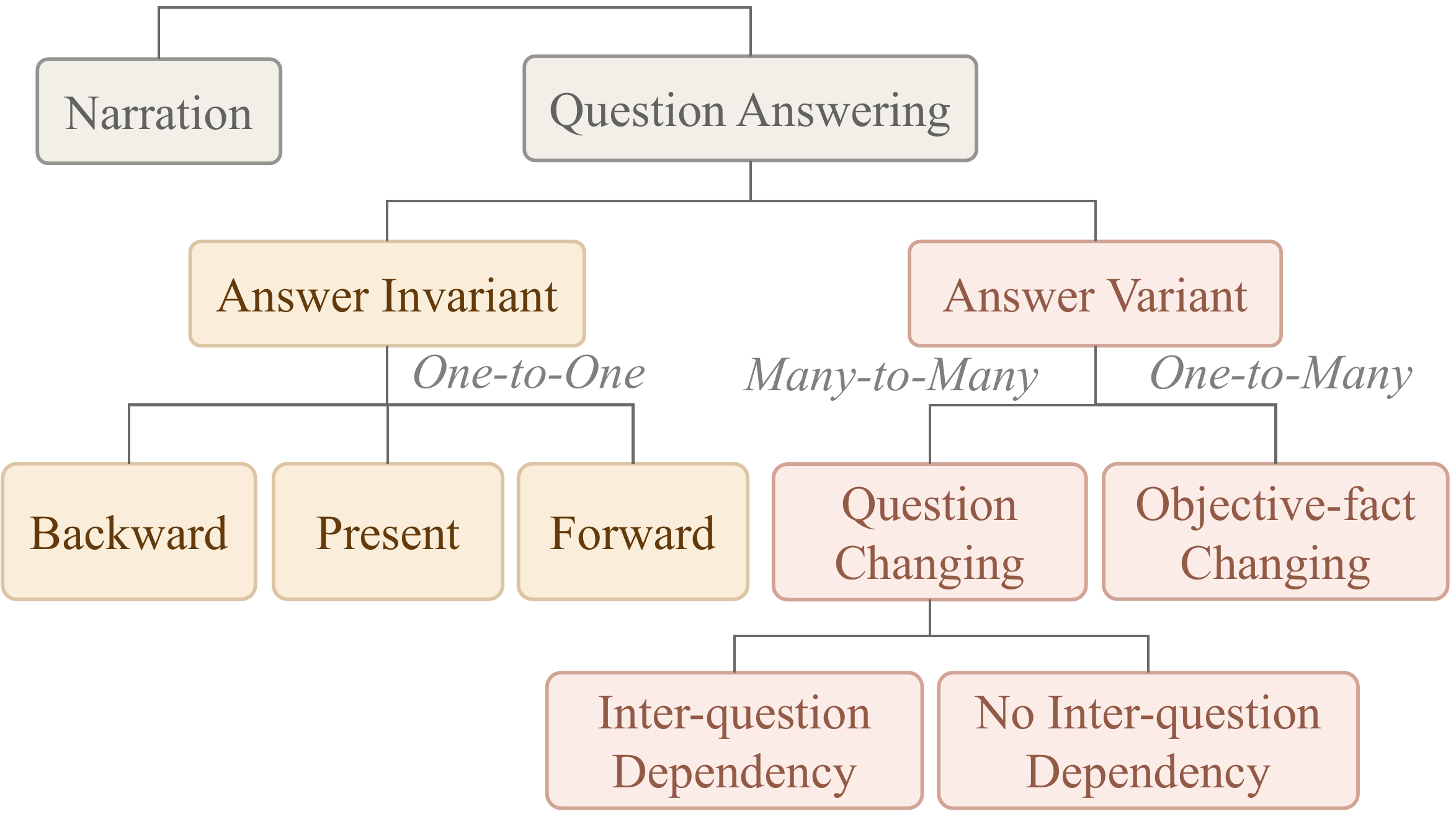}
    \caption{Overview of our task taxonomy.}
    \label{fig:taxonomy}
\end{wrapfigure}
We integrate multiple open-source video datasets as sources, including EpicKitchens~\cite{damen2020epickitchensdatasetcollectionchallenges},
YouCook2~\cite{zhou2017automaticlearningproceduresweb}, Ego4D~\cite{9879279}, 
EgoBlind~\cite{xiao2025egoblindegocentricvisualassistance}, EgoExo4D~\cite{grauman2024egoexo4dunderstandingskilledhuman}, EgoExoLearn~\cite{huang2025egoexolearndatasetbridgingasynchronous},   DiDeMo~\cite{8237880}, Charades~\cite{sigurdsson2018charadesegolargescaledatasetpaired}, OmniStar~\cite{yang2025livestar}, QVHighlights~\cite{10205244}, ActivityNet~\cite{7298698}, Shot2Story~\cite{han2025shot2storynewbenchmarkcomprehensive},  Assembly101~\cite{9878834}, HoloAssist~\cite{wang2023holoassistegocentrichumaninteraction}, WTaG~\cite{bao-etal-2023-foundation}, MovieChat~\cite{10657734}, QueryD~\cite{9414640}, and Live-WhisperX~\cite{11092811}.
To convert the raw dataset into the standard multi-turn online interaction format (see Section~\ref{sec:3}), we perform per-second frame-text alignment, in which each one-second segment is paired with either an explicit response or a silence token and tagged with a timestamp (e.g., \texttt{<0s-1s>}).
In addition, we conduct source-level inspection and discard sources with uncertain annotations.
This yields 248K samples spanning diverse real-world scenarios such as \textit{daily living}, \textit{work and productivity}, \textit{knowledge and learning}, and \textit{outdoor and recreation}. 

\subsection{Task Taxonomy}

Figure~\ref{fig:taxonomy} provides an overview of our task taxonomy.
The first, \textbf{narration}, requires the model to continuously monitor the incoming video and produce coherent descriptions of salient events.
We build this task from dense temporal annotations and automatic speech recognition (ASR) transcripts. Rather than captioning every frame, the model emits a description only when such events occur.
The second, \textbf{question answering}, requires the model to respond selectively conditioned on the observed video frames. 
We categorize QA by whether the answer varies across observations: \textit{(1) Answer-invariant}, a one-to-one question–answer mapping; and \textit{(2) Answer-variant}, further partitioned into \textit{(2.1) Objective-fact changing}, a one-to-many mapping, and \textit{(2.2) Question changing}, a many-to-many mapping. \textit{Question changing} is further divided into \textit{(2.2.1) Inter-question dependency} and \textit{(2.2.2) No inter-question dependency}.
Orthogonally, the evidence frames may be situated \textit{backward}, \textit{present}, or \textit{forward} relative to the query.

\subsection{Training Objective}
In streaming scenarios, silence runs are substantially more frequent than response segments, so the token distribution over assistant positions is heavily skewed toward \texttt{</silence>}. Under standard supervised fine-tuning, this imbalance biases the gradient toward reinforcing continued silence rather than the transition from silence to response, diluting the response signal. To address this, we assign distinct weights to tokens based on their role. Let $\mathcal{A}$ denote the set of supervised assistant-token positions, and let $\mathcal{C}=\{c_i\}_{i=1}^{M}\subseteq\mathcal{A}$ denote the ordered subsequence of control-token positions, where $y_{c_i}\in\{\texttt{</silence>}, \texttt{</response>}\}$. A \textit{silence run} is a maximal consecutive subsequence of \texttt{</silence>} tokens within $\mathcal{C}$. For $j\in\mathcal{A}$, with the convention $y_{c_0}=\varnothing$, we assign:
\begin{equation}
w_j =
\begin{cases}
w^{\text{first}}_{\text{silence}}, & y_{c_i}=\texttt{</silence>},\ y_{c_{i-1}}\neq\texttt{</silence>}, \\
w^{\text{repeated}}_{\text{silence}}, & y_{c_i}=\texttt{</silence>},\ y_{c_{i-1}}=\texttt{</silence>}, \\
w_{\text{response}}, & y_{c_i}=\texttt{</response>}, \\
1, & \text{otherwise.}
\end{cases}
\label{eq:token_weights}
\end{equation}
We set $w^{\text{first}}_{\text{silence}}=1$, $w^{\text{repeated}}_{\text{silence}}<1$ to down-weight silence continuations, and $w_{\text{response}}>1$ to up-weight response onsets. The training objective normalizes by the number of supervised  positions:
\begin{equation}
\mathcal{L}(\theta) = -\frac{1}{|\mathcal{A}|}\sum_{j\in\mathcal{A}} w_j \, \log p_\theta\!\left(y_j \mid y_{<j}\right).
\label{eq:final_loss}
\end{equation}

\section{Streaming Harness}
\label{sec:5}
While the Streaming-Native VLM trained in Section~\ref{sec:4} acquires the ability to emit \texttt{</silence>} and \texttt{</response>} appropriately, deploying it on open-ended streams uncovers another class of challenges that lie beyond model capability and instead stem from the inference pipeline itself. 
Real-world streams last for hours rather than the minutes typical of training clips, demanding a memory mechanism whose footprint does not grow without bound as stream length increases.
Moreover, to keep input frames and output text synchronized, each step must also be processed within a sub-second budget to preserve the responsiveness on which proactive interaction depends.

A straightforward remedy is to maintain the KV cache via a sliding window~\cite{xu2026streamingvlm,shen2026simplebaselinestreamingvideo}; an orthogonal line of work instead prunes redundant tokens to curtail cache growth~\cite{yao2025timechatonline}. 
While both strategies effectively alleviate memory overhead, neither supports KV cache reuse, precluding direct integration with the prefix-caching optimizations exploited by modern inference engines~\cite{kwon2023efficientmemorymanagementlarge}.

\begin{figure}
  \centering
  \includegraphics[width=1\textwidth]{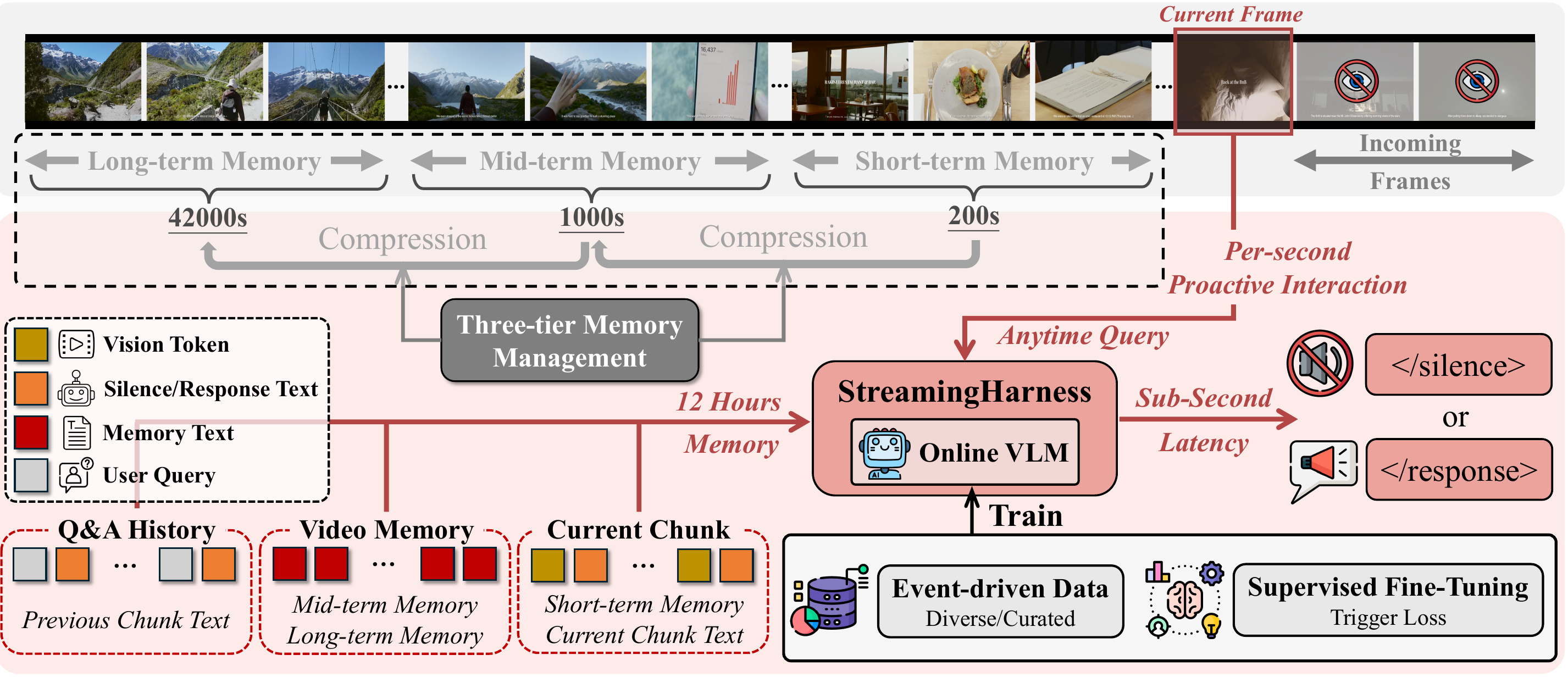} 
  \caption{\textbf{Overview of StreamingHarness.} A three-tier memory management scheme compresses the streaming video into short-, mid-, and long-term memory (200 s / 1{,}000 s / 42{,}000 s), enabling up to \textbf{12 hours} of context. At each step, the Online VLM decides whether to emit \texttt{</silence>} or \texttt{</response>} with sub-second latency, enabled by our \texttt{vLLM}-friendly design.}
  \label{fig:method}
\end{figure}

To address these challenges and enable proactive interaction, we introduce StreamingHarness, a plug-and-play inference framework tailored to real-world streaming video applications. StreamingHarness supports unbounded streaming video inference with hours-long memory retention under stringent real-time latency constraints. The overall architecture is illustrated in Figure~\ref{fig:method}.
Specifically, StreamingHarness comprises three components: \textit{three-tier memory management}, \textit{prefix-aware KV cache}, and \textit{event-driven response triggering}.
  
\subsection{Three-tier Memory Management}
\label{sec:5.1}
To accommodate hours-long streams within a bounded memory budget, we organize the context into a three-tier hierarchy. As shown in Figure~\ref{fig:method}, the stream is partitioned into: (i) a \emph{short-term memory} retaining the most recent $T_s$ seconds as raw vision tokens; (ii) a \emph{mid-term memory} holding up to $M$ textual summaries of past short-term chunks, covering $T_m = M T_s$ seconds at moderate compression; and (iii) a \emph{long-term memory} storing up to $L$ aggressively compressed blocks, each consolidated from $M$ consecutive mid-term summaries and thus spanning $T_l = L M T_s$ seconds.

\begin{wrapfigure}{r}{0.55\textwidth}  
    \centering
    \includegraphics[width=0.55\textwidth]{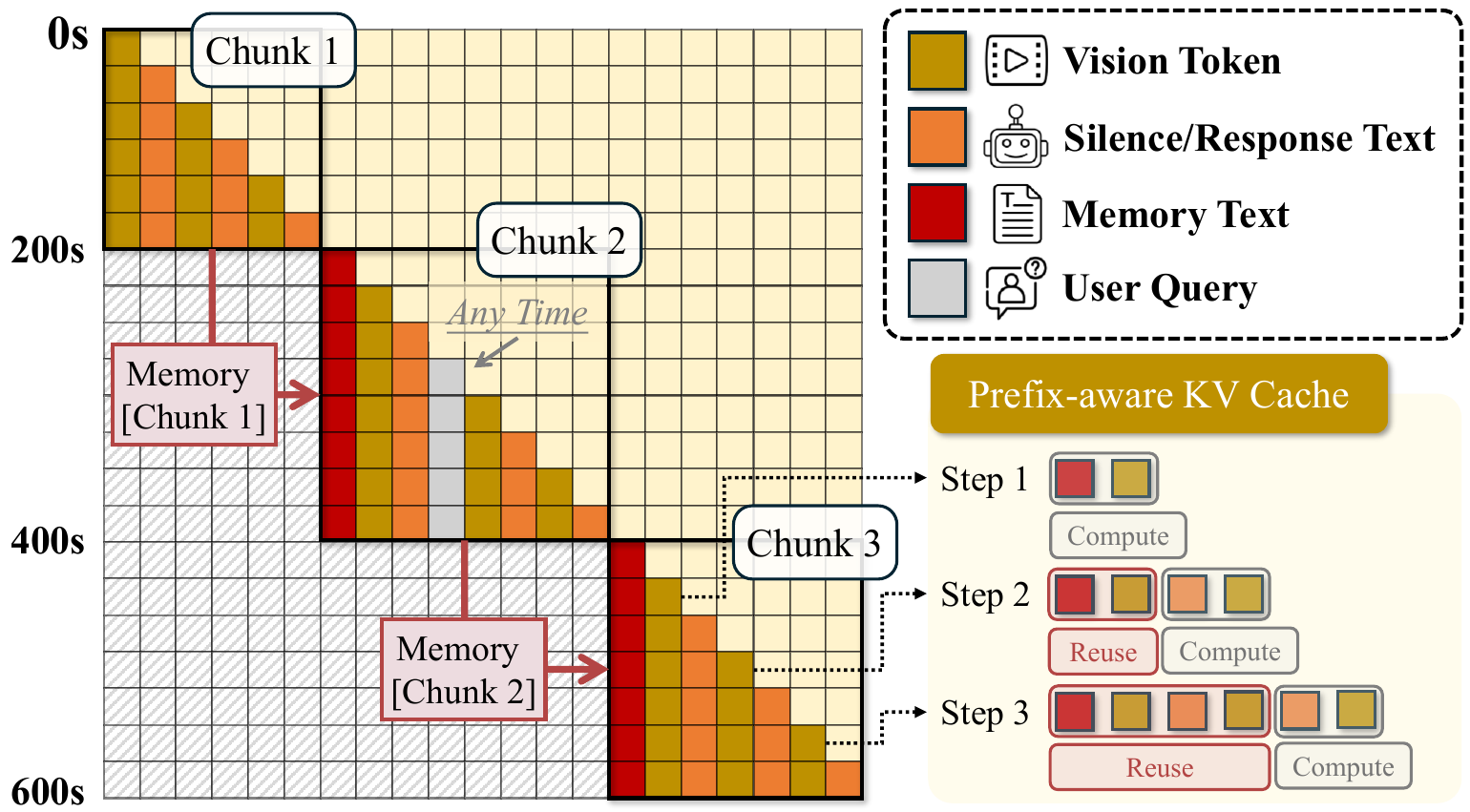}
    \caption{At each chunk, StreamingHarness reuses the KV cache.}
    \label{fig:attention}
\end{wrapfigure}
Concretely, whenever the short-term buffer fills, the enclosed chunk is offloaded through a \textit{mid-term memory agent}, which condenses $T_s$ seconds of frames into a compact textual summary that faithfully captures per-frame semantics, salient details, and the temporal evolution of object and scene states, trading vision tokens for text tokens at substantially lower cost. Once $M$ such summaries accumulate, a \textit{long-term memory agent} consolidates them into a single block by merging overlapping content, removing redundancy, and organizing causally related events into a coherent timeline. The long-term tier is capped at $L$ blocks with FIFO eviction, and both agents run asynchronously ahead of each chunk boundary, fully hiding memory consolidation behind mainline inference. The total temporal coverage is therefore bounded by
$T_s + T_m + T_l \;=\; T_s\,(1 + M + L M),$
which, under our default configuration ($T_s{=}200$\,s, $M{=}5$, $L{=}42$), reaches roughly 12 hours, sufficient for the vast majority of real-world streams.

\subsection{Prefix-aware KV Cache}
\label{sec:prefix}

Although the three-tier hierarchy already shortens the context, recomputing the entire KV cache at every step's prefill stage would still violate the sub-second latency budget. Modern inference engines~\cite{kwon2023efficientmemorymanagementlarge,zheng2024sglangefficientexecutionstructured} accelerate the prefill stage via \emph{prefix caching}, which caches the KV states of prior dialogue turns and reuses them whenever a subsequent request shares the same prefix. StreamingHarness is therefore organized around this principle.

As illustrated in Figure~\ref{fig:attention}, StreamingHarness directs the engine to perform a one-time prefill of the Memory Text at the start of each chunk, converting it into the corresponding KV cache. At every subsequent step, the new frame and the previous response are appended, and only their KV entries need to be computed, while all previously cached KV states, including the Memory Text and earlier turns within the chunk, are reused without recomputation. This design substantially reduces the per-step time-to-first-token (TTFT), enabling StreamingHarness to deliver near-instantaneous responses.

\subsection{Event-driven Response Triggering}

We propose a training-free adaptation scheme that enables a base model without native proactive-interaction capability to generalize to proactive-interaction objectives. At each step, the model is conditioned on the current frame, the dialogue context of the current chunk, the past video history, and the past question-answer history. Given the active query, the model performs a judgment over this evidence: if no new and informative content is yet available, for instance when the key event has not occurred or the scene is still in transition, it remains silent and defers to the next frame; otherwise, it produces a concise response strictly grounded in the visible evidence, without fabricating unseen entities or facts. 
The corresponding system prompt is provided in Appendix~\ref{appendix:prompt}.
This procedure yields a streaming inference loop of four stages: query triggering, frame-by-frame judgment, timely response, and monitoring resumption.

Together, these designs endow StreamingHarness with the long-horizon memory, real-time inference, and proactive interaction required for production-grade streaming video understanding.

\section{Streaming-Eval}
To facilitate community progress on real-time streaming capabilities, we introduce Streaming-Eval, a benchmark specifically designed for in-the-wild streaming video scenarios, which jointly assesses both answer correctness and response timeliness.
\subsection{Benchmark Construction}

\begin{figure}
  \centering
  \includegraphics[width=1\textwidth]{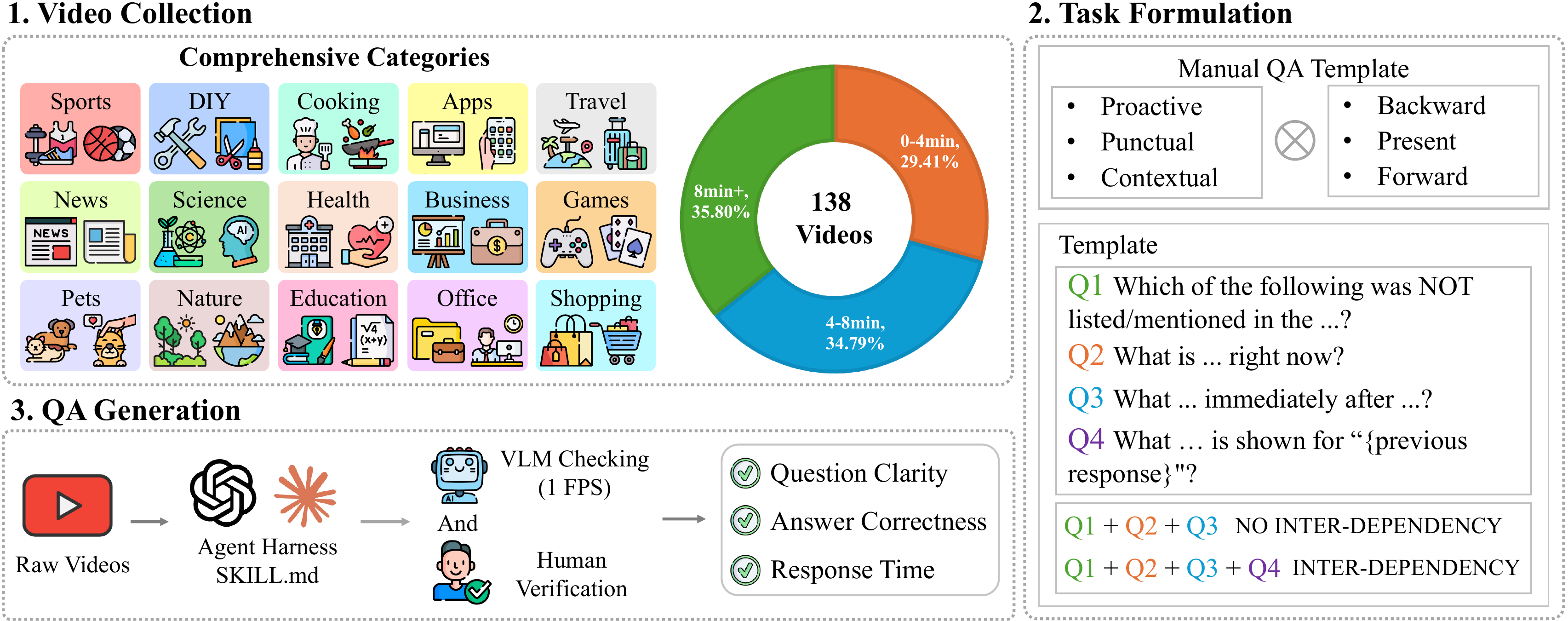} 
  \caption{\textbf{Generation pipeline of Streaming-Eval.} We curate high-quality and diverse public videos, and automatically generate multiple-choice questions through an agent harness guided by a well-designed system prompt. GPT 5.4 and Claude Opus 4.6 are adopted as the VLMs for annotation.}
  \label{fig:data}
\end{figure}

Figure~\ref{fig:data} illustrates the three-stage construction pipeline of Streaming-Eval: video collection, task formulation, and QA generation.
\textbf{\textit{(i) Video collection.}} We aggregate 138 videos from PhoStream~\cite{lu2026phostreambenchmarkingrealworldstreaming}, Video-MMMU~\cite{hu2025videommmuevaluatingknowledgeacquisition}, Ego4D~\cite{9879279}, THUMOS~\cite{Idrees_2017}, HiREST~\cite{zala2023hierarchicalvideomomentretrievalstepcaptioning}, MLVU~\cite{zhou2025mlvubenchmarkingmultitasklong}, RTV-Bench~\cite{xun2026rtvbench}, and OVO-Bench~\cite{11094230}, covering 15 everyday categories with a balanced split across short (0--4 min), medium (4--8 min), and long (8 min+) clips.
\textbf{\textit{(ii) Task formulation.}} To systematically evaluate streaming video comprehension, we adopt the same task taxonomy as that used for the training data, summarized in Figure~\ref{fig:taxonomy}. Each task type is constructed from a carefully designed template that pairs one mode of streaming interaction (\textit{proactive}, \textit{punctual}, or \textit{contextual}) with one mode of streaming understanding (\textit{backward}, \textit{present}, or \textit{forward}), thereby yielding comprehensive coverage of streaming interaction patterns.
\textbf{\textit{(iii) QA generation.}} Both questions and answers are generated by an agent harness guided by a structured \textsc{Skill}, with GPT 5.4 and Claude Opus 4.6 serving as the annotation models. Each QA pair is then filtered by a VLM checker and verified by human annotators along three dimensions: question clarity, answer correctness, and response time.

\subsection{Evaluation Metric}
For narration, we sample 100 instances from the LiveSports3K-CC benchmark~\cite{11092811}. For each sample, we convert each candidate's narration into a timestamped transcript of the form \texttt{[$t$\,s] utterance}, where $t$ denotes the time at which the utterance is emitted, and prompt Claude Sonnet~4.6 to compare the two candidates against the ground-truth Automatic Speech Recognition (ASR) transcription. The judge evaluates two criteria of equal weight: narration style (pacing and tone) and consistency with the reference. To mitigate position bias, each pair is judged twice with the candidate order swapped, and we report the win rate over the resulting $2N$ votes. The full judge prompt is provided in Appendix~\ref{appendix:narration}.

For question answering, we use a rule-based scheme. Each question is annotated with a ground-truth answer window $W$. A prediction is a true positive (TP) if it is emitted within $W$ and exactly matches the ground truth; any non-silence prediction outside $W$ is a false positive (FP), and a question with no valid prediction in $W$ is a false negative (FN).
To jointly assess correctness and timeliness in the streaming setting while treating answer correctness as our primary concern, we adopt a streaming weighted F1 (SW-F1) that assigns higher weights to TP and FN:
\begin{equation}
    \text{SW-F1} = \frac{w_{\text{TP}}\,\mathrm{TP}}{w_{\text{TP}}\,\mathrm{TP} + w_{\text{FP}}\,\mathrm{FP} + w_{\text{FN}}\,\mathrm{FN}},
\end{equation}
with $w_{\text{TP}}=w_{\text{FN}}=2.0$ and $w_{\text{FP}}=0.2$.
Compared to accuracy-style metrics that count only TP, the FP term penalizes "always-respond" behaviors that emit answers at every step to inflate TP. Compared to vanilla F1, the lower FP weight prioritizes answer correctness, preventing over-penalization.

\section{Experiment}

\subsection{Experimental Setup}
\label{sec:7.1}

\paragraph{Training.}

We continue training from Qwen3-VL-8B-Instruct~\cite{bai2025qwen3vltechnicalreport} on Streaming-Train-248K with a learning rate of $1\!\times\!10^{-5}$. During fine-tuning, the vision encoder is kept frozen, while the connector and the LLM are fully updated. Each training video is sampled at 1\,FPS, with the maximum number of pixels per frame capped at 131{,}072 and the maximum context length set to 48K tokens. For the token weights in Eq.~\ref{eq:token_weights}, we set $w_{\text{silence}}^{\text{first}}\!=\!1$, $w_{\text{silence}}^{\text{repeated}}\!=\!0.8$, and $w_{\text{response}}\!=\!1.5$. The full training run consumes approximately 4{,}096 H200 GPU-hours.

\paragraph{Baselines.}
We evaluate two categories of models as baselines in this study. (1) Top closed-source models, including Claude Opus 4.6 Thinking, GPT 5.4 (High), Gemini 3.1 Pro (High), and Doubao Seed 2.0 Pro. (2) Open-source models, including Qwen3-VL-8B, Qwen3-VL-32B, and Qwen3.5-9B.

\paragraph{Inference.}
All baselines process videos at $1$\,FPS, with the maximum number of pixels per frame capped at $262{,}144$.
For our three-tier memory management (Sec.~\ref{sec:5.1}), we adopt the default configuration $T_s\!=\!200$\,s, $M\!=\!5$, $L\!=\!42$ for open-source models. For closed-source models, we reduce the configuration to $T_s\!=\!10$\,s, $M\!=\!4$, $L\!=\!8$ due to API rate limits.
The mid-term memory agent is capped at $262{,}144$ input pixels per frame and $4{,}000$ output tokens, while the long-term memory agent is capped at $2{,}000$ output tokens.
\subsection{Accuracy Evaluation}

\newcommand{\perfcellthree}[3][]{%
\begin{tikzpicture}[baseline=(bar.base)]
    \pgfmathsetmacro{\W}{1.4}
    \pgfmathsetmacro{\H}{0.25}
    \pgfmathsetmacro{\cutA}{#2}
    \pgfmathsetmacro{\cutB}{100 - #3}
    \fill[teal!40]              (0,0)                 rectangle ({\W*\cutA/100},\H);
    \fill[gray!40]              ({\W*\cutA/100},0)    rectangle ({\W*\cutB/100},\H);
    \fill[red!40]               ({\W*\cutB/100},0)    rectangle (\W,\H);
    \ifthenelse{\equal{#1}{false}}{}{%
        \node[font=\small, text=black, anchor=west] at (0.03,{\H/2}) 
            {#1{#2}};%
    }%
    \node[inner sep=0pt, outer sep=0pt, anchor=base] (bar) at (0,-0.05ex) {};
\end{tikzpicture}%
}
\begin{table}[t]
\centering
\caption{\textbf{Narration accuracy (win rate vs.\ baselines).} $^\dagger$ denotes offline models adapted to our streaming harness framework. The colored bars (e.g., \scalebox{0.6}{\perfcellthree[false]{33.0}{33.0}}) indicate the \textcolor{teal!40}{win} / \textcolor{gray}{tie} / \textcolor{red!40}{loss} rates (\%) against the baseline, with the displayed number denoting the \textcolor{teal!40}{win} rate. Due to API rate limits, $T_s$ is set to 10 for closed-source models.}
\label{tab:narration}
\renewcommand{\arraystretch}{1} 
\setlength{\tabcolsep}{5.0pt}
\begin{tabular}{l|cccc|c}
\toprule
\multirow{3}{*}{\textbf{Method}} & \multicolumn{4}{c|}{\textbf{Narration (Win Rate \%)}} & \multirow{3}{*}{\textbf{Avg.}} \\
\cmidrule(r){2-5}
 & {\footnotesize \makecell{vs. Claude Opus \\ 4.6 Thinking}} & {\footnotesize \makecell{vs. GPT \\ 5.4 (High) }} & {\footnotesize \makecell{vs. Gemini \\3.1 Pro (High)}} &  {\footnotesize \makecell{vs. Doubao \\Seed 2.0 Pro}} & \\
\midrule
Claude Opus 4.6 Thinking  $^\dagger$      & \perfcellthree{14.5}{6.5}  & \perfcellthree{10.5}{50.0}  & \perfcellthree{10.0}{72.5} & \perfcellthree{26.0}{26.0} & 15.3 \\
GPT 5.4 (High) $^\dagger$                 & \perfcellthree{52.5}{19.5} & \perfcellthree{12.5}{12.5}   & \perfcellthree{18.0}{59.0} & \perfcellthree{44.0}{18.5} & 31.8 \\
Gemini 3.1 Pro (High) $^\dagger$          & \perfcellthree[\textbf]{70.0}{7.0} & \perfcellthree{62.0}{10.5} & \perfcellthree{35.5}{24.0}   & \perfcellthree{69.0}{6.0}   & 59.1 \\
Doubao Seed 2.0 Pro $^\dagger$            & \perfcellthree{19.5}{28.5} & \perfcellthree{11.5}{47.0} & \perfcellthree{7.0}{66.0}   & \perfcellthree{10.0}{13.5}    & 12.0 \\
\midrule

Qwen3-VL-8B    $^\dagger$                 & \perfcellthree{4.0}{90.5}  & \perfcellthree{4.0}{92.5}   & \perfcellthree{2.5}{94.5}   &  \perfcellthree{4.0}{87.0}   & 3.6  \\
Qwen3-VL-32B   $^\dagger$                 & \perfcellthree{3.0}{90.5}   & \perfcellthree{3.0}{94.0}   & \perfcellthree{0.5}{98.0}  & \perfcellthree{2.0}{91.0}  & 2.1  \\
Qwen3.5-9B (Thinking) $^\dagger$          & \perfcellthree{0.5}{99.0}   & \perfcellthree{0.5}{98.0}   & \perfcellthree{0.0}{100.0}   & \perfcellthree{1.0}{98.5}  & 0.5  \\
\midrule
StreamingHarness-8B $^\dagger$             & \perfcellthree{65.0}{29.0} & \perfcellthree[\textbf]{65.0}{26.0} & \perfcellthree[\textbf]{44.0}{45.0}  & \perfcellthree[\textbf]{71.5}{24.0} & \textbf{61.4} \\
\bottomrule
\end{tabular}
\end{table}

\paragraph{Narration}

\begin{figure}
  \centering
  \includegraphics[width=0.8\textwidth]{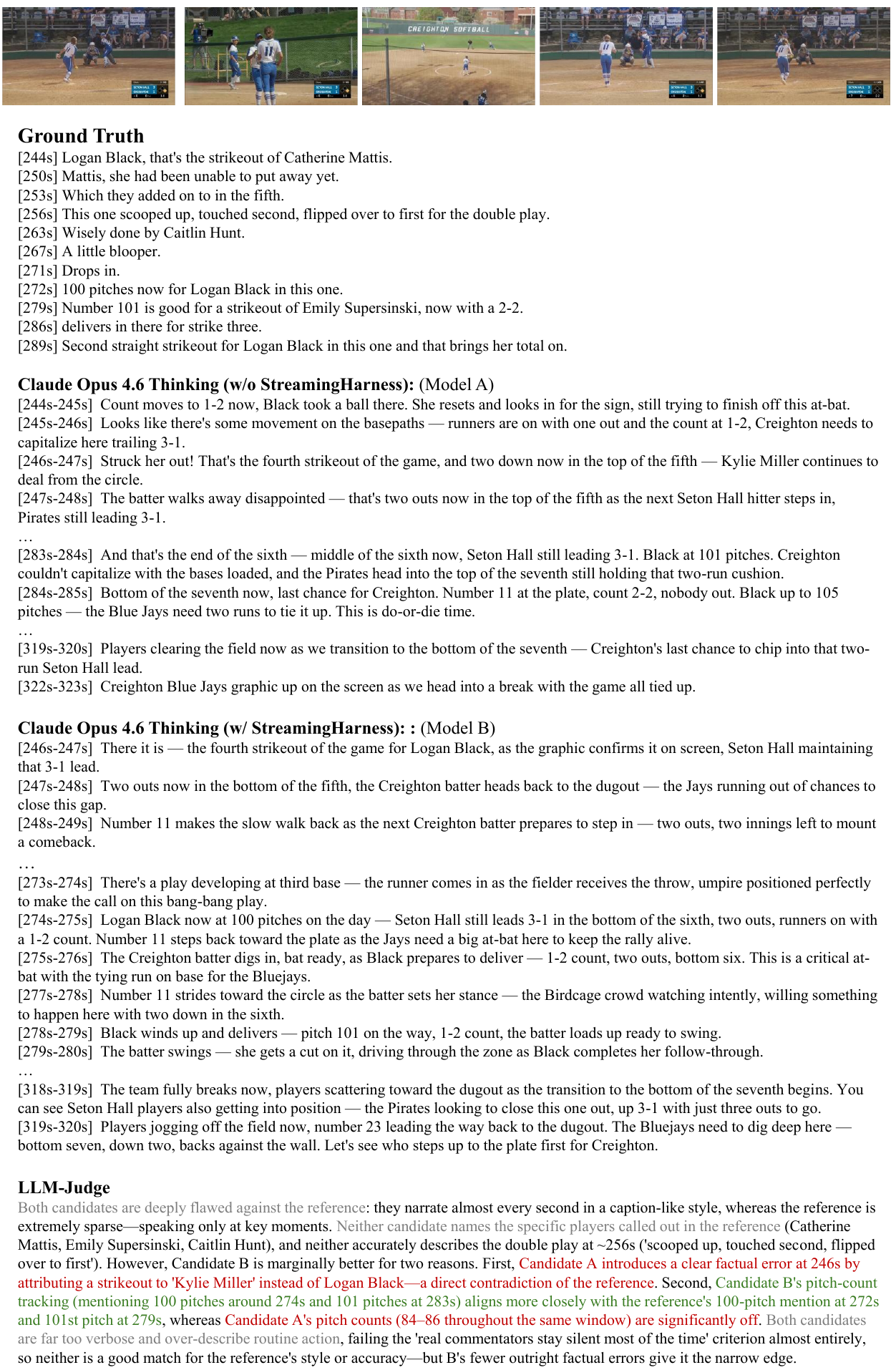} 
  \caption{Narration comparison between Claude Opus 4.6 Thinking (w/o StreamingHarness) and Claude Opus 4.6 Thinking (w/ StreamingHarness). }
  \label{fig:case2}
\end{figure}

\begin{figure}
  \centering
  \includegraphics[width=0.9\textwidth]{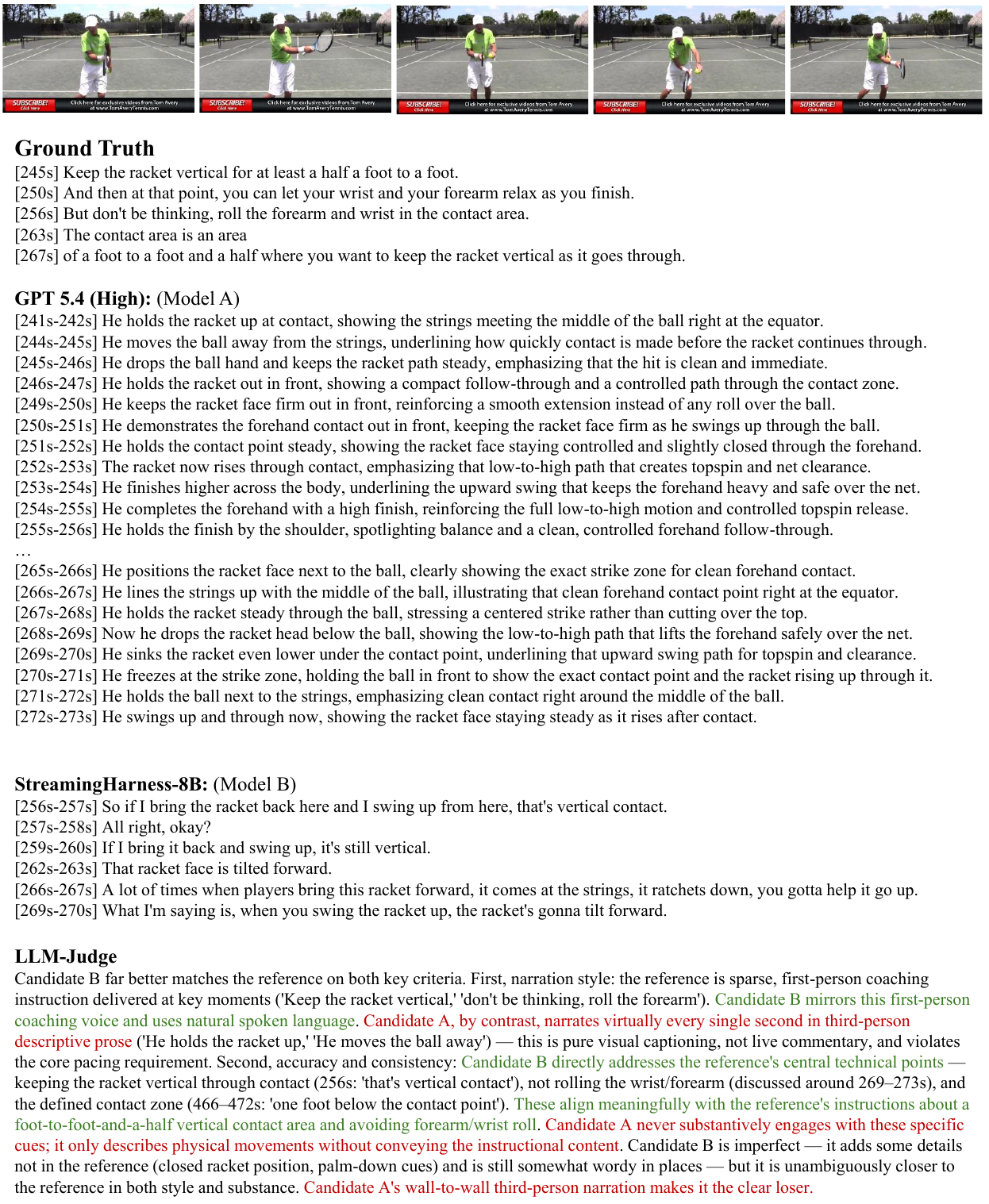} 
  \caption{Narration comparison between GPT 5.4 (High) and StreamingHarness-8B. }
  \label{fig:case1}
\end{figure}

We evaluate narration accuracy by reporting pairwise win rates against four closed-source VLMs, as shown in Table~\ref{tab:narration}.
We find that both closed-source and open-source baselines over-narrate in a dense captioning style, failing to produce natural commentary that speaks only at key moments.
Smaller open-source models perform particularly poorly, as they frequently hallucinate, misreport critical outcomes, repeat templated phrases, and miss key events.
In contrast, our 8B streaming-native VLM achieves a $61.4\%$ average win rate, reaching $71.5\%$ against Doubao Seed 2.0 Pro and $65.0\%$ against both Claude Opus 4.6 Thinking and GPT 5.4 (High). 
The reason is that large-scale training on in-the-wild tasks, paired with a tailored loss function (Section~\ref{sec:4}), gives the model proactive interaction capabilities and lets it produce output that reads like live broadcast commentary (see Figure~\ref{fig:case1}).
We further observe that Claude Opus 4.6 Thinking$^\dagger$ and Gemini 3.1 Pro$^\dagger$ (High) achieve higher win rates than loss rates against Claude Opus 4.6 Thinking and Gemini 3.1 Pro (High), respectively. The improvement stems from our three-tier memory management (Section~\ref{sec:5.1}), which retains context such as player names and scores to anchor subsequent commentary and mitigate hallucination (see Figure~\ref{fig:case2}).

\paragraph{Question Answering}

\begin{figure}
  \centering
  \includegraphics[width=0.7\textwidth]{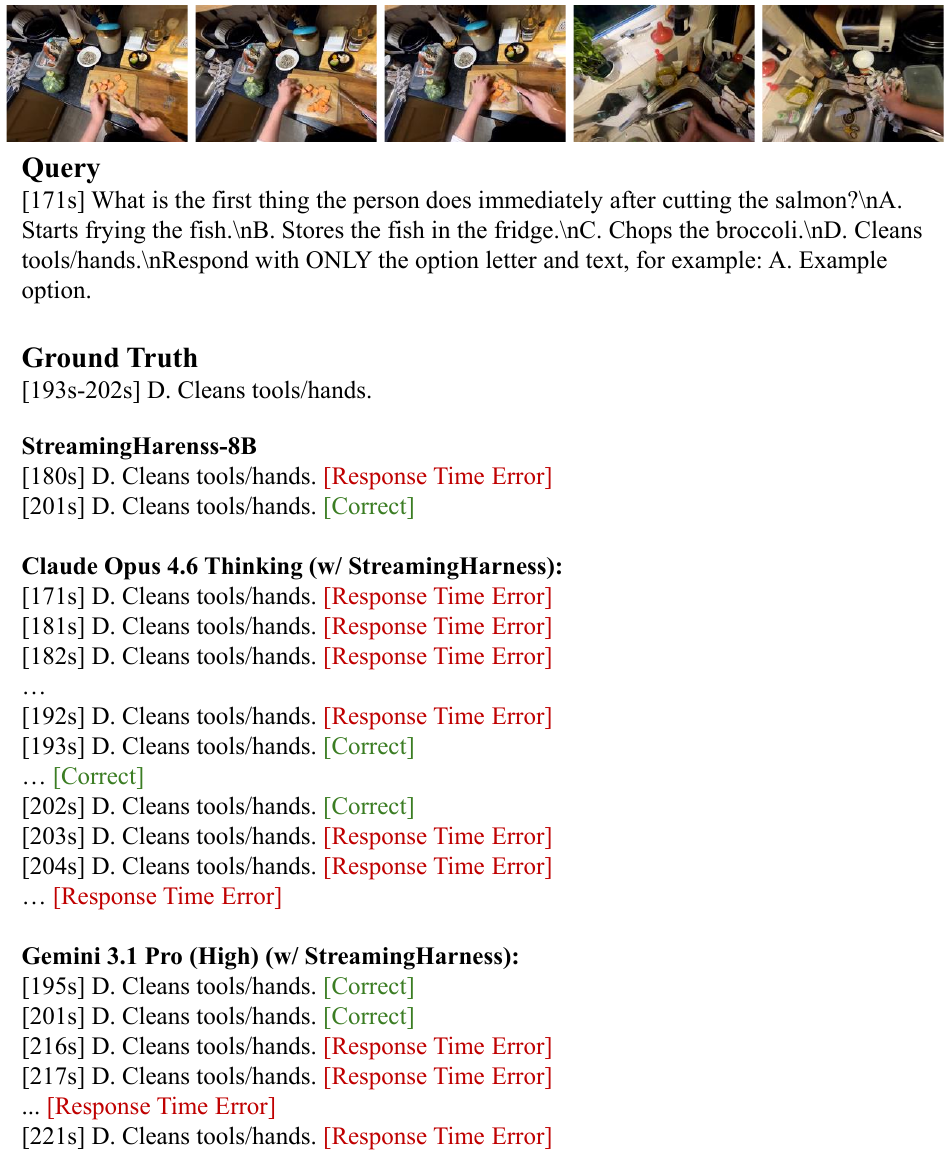} 
  \caption{Forward Task Example. }
  \label{fig:case3}
\end{figure}

As shown in Table~\ref{tab:qa}, we conduct a systematic comparison between StreamingHarness-8B and the baselines on QA tasks. The results demonstrate that StreamingHarness-8B achieves significant improvements over all current VLMs across all streaming QA tasks.
Specifically, our harness (Section~\ref{sec:5}) retrieves past evidence from the memory text and preserves Q\&A history across turns, yielding large gains on Backward and Inter-question Dependency (see Figure~\ref{fig:case4}). Meanwhile, our 8B streaming-native VLM (Section~\ref{sec:4}) enables stronger proactive interaction by deferring responses until future evidence becomes available (see figure~\ref{fig:case3}), substantially improving Forward and Objective-fact Changing.

\begin{table}[t]
\centering
\caption{\textbf{Evaluation results of different models on Streaming-Eval (QA).} Scores reported in the table are $\text{SW-F1}$. $^\dagger$ denotes offline models adapted to our streaming harness framework. Due to API rate limits, $T_s$ is set to 10 for closed-source models.}
\label{tab:qa}
\setlength{\tabcolsep}{5pt}
\renewcommand{\arraystretch}{1} 
\begin{tabular}{l|ccc|rrc|c}
\toprule
\multirow{3}{*}{\textbf{Method}}
& \multicolumn{3}{c|}{\textbf{Answer Invariant QA}} 
& \multicolumn{3}{c|}{\textbf{Answer Variant QA}} 
& \multirow{3}{*}{\textbf{Avg.}} \\
\cmidrule(lr){2-4} \cmidrule(lr){5-7}
& {\small Backward} & {\small Present} & {\small Forward} & {\small OFC} & {\small OC \tiny w/ IQD} & {\small OC \tiny w/o IQD} & \\

\midrule

Claude Opus 4.6 Thinking $^\dagger$  & 12.5 & 6.1  & 10.3 & 8.6  & 11.1 & 8.4  & 9.5  \\
GPT 5.4 (High) $^\dagger$            & 26.0 & 17.4 & 15.1 & 9.9  & 20.2 & 21.0 & 18.3 \\
Gemini 3.1 Pro (High)  $^\dagger$    & 17.0 & 8.2  & 16.9 & 11.9 & 14.7 & 14.2 & 13.8 \\
Doubao Seed 2.0 Pro   $^\dagger$     & 14.6 & 9.7  & 17.9 & 13.2 & 13.1 & 11.3 & 13.3 \\
\midrule
Qwen3-VL-8B     $^\dagger$           & 16.2 & 5.7 & 13.7 & 7.0 & 10.2 & 12.0 & 10.8 \\
Qwen3-VL-32B     $^\dagger$          & 17.0 & 8.0 & 15.2 & 6.8 & 10.4 & 13.8 & 11.9 \\
Qwen3.5-VL-9B (Thinking)  $^\dagger$ & 0.0  & 0.0 & 0.1  & 0.0 & 0.2  & 0.3  & 0.1  \\
\midrule
StreamingHarness-8B $^\dagger$        & \textbf{54.2} & \textbf{21.0} & \textbf{48.9} & \textbf{32.5} & \textbf{61.9} & \textbf{56.3} & \textbf{45.8} \\
\bottomrule
\end{tabular}
\end{table}
\subsection{Efficiency Evaluation}
\begin{figure}
  \centering
  \includegraphics[width=1\textwidth]{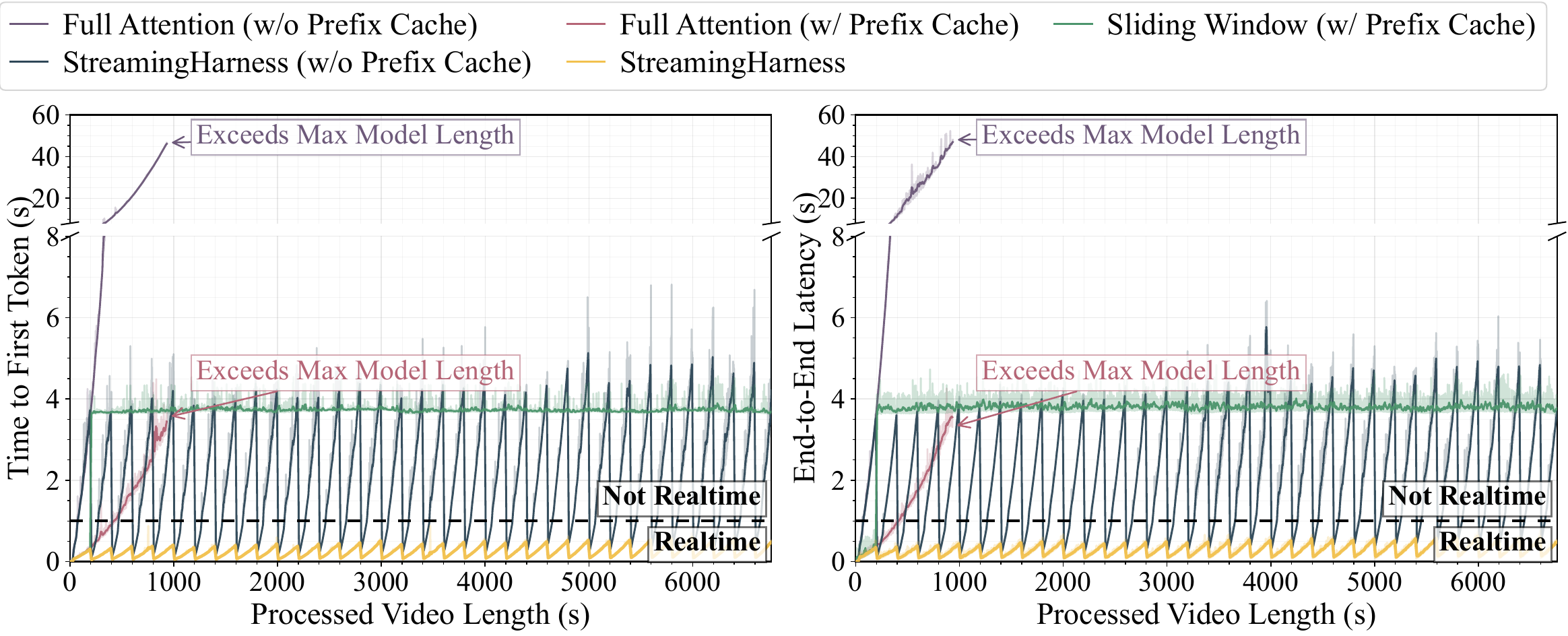} 
  \caption{Time to First Token (left) and End-to-End Latency (right) vs. processed video length (up to 2 h). StreamingHarness keeps latency low and stable, below the real-time threshold.}
  \label{fig:latency}
\end{figure}

As shown in Figure~\ref{fig:latency}, we evaluate TTFT and End-to-End latency for five inference strategies on a 2-hour sports broadcast video~\cite{xu2026streamingvlm}: full attention (w/o prefix cache), full attention (w/ prefix cache), sliding window (w/ prefix cache), StreamingHarness (w/o prefix cache), and our full StreamingHarness. All measurements are conducted at 1 FPS on a single NVIDIA H200. The real-time threshold is set to 1 s, matching the inter-frame interval and ensuring that each response is generated before the next input arrives.

Full attention saturates the model's maximum context length within a few hundred seconds, regardless of prefix caching.
Sliding window prevents context overflow and effectively alleviates memory overhead, but its latency stabilizes at approximately 4 s, substantially exceeding the real-time threshold; since consecutive steps share no common prompt prefix, \texttt{vLLM}'s~\cite{kwon2023efficientmemorymanagementlarge} prefix caching mechanism provides no benefit under this scheme.
In contrast, our full StreamingHarness is natively compatible with \texttt{vLLM}'s prefix caching, as detailed in Section~\ref{sec:prefix}. Compared with StreamingHarness (w/o prefix cache), enabling prefix caching reduces latency from approximately 4 s to below 1 s. As a result, both TTFT and End-to-End latency of StreamingHarness remain consistently below the real-time threshold throughout the entire 2-hour video.

\subsection{Ablation Study}

\begin{wrapfigure}{r}{0.6\textwidth}  
    \centering
    \includegraphics[width=0.6\textwidth]{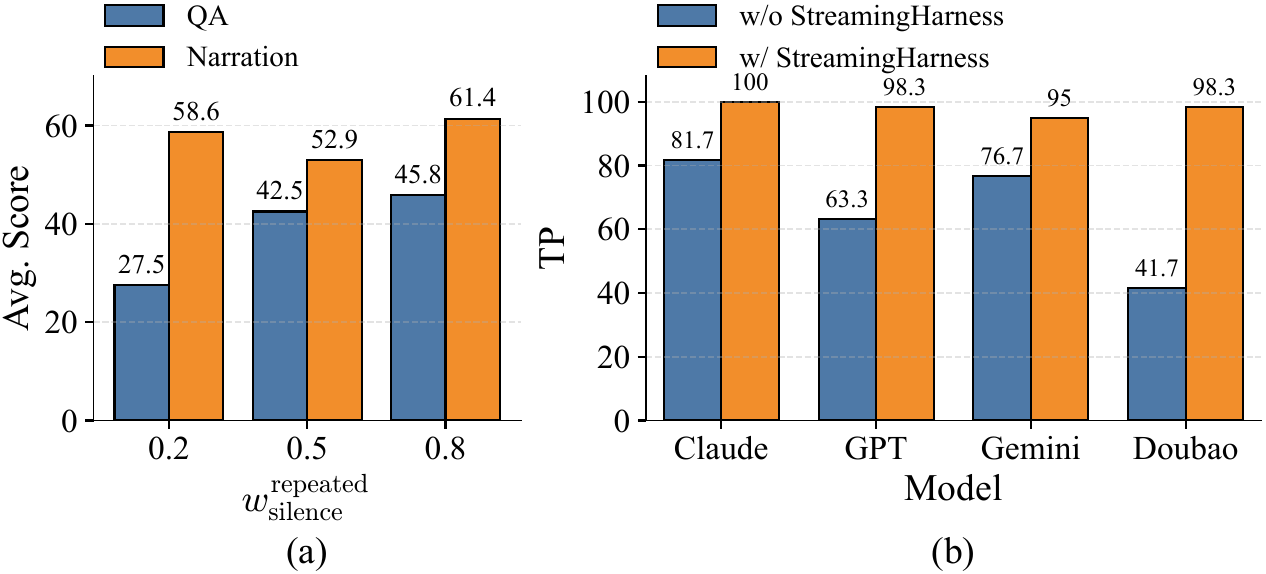}
    \caption{\textbf{Ablation studies.} (a) Avg.\ SW-F1 on QA and Avg.\ win rate on Narration under varying repeated-silence weights $w_{\text{silence}}^{\text{repeated}} \in \{0.2,\, 0.5,\, 0.8\}$ in Eq.~(\ref{eq:token_weights}). (b) Effect of StreamingHarness on the Backward task. }
    \label{fig:ablation}
\end{wrapfigure}

To evaluate the effectiveness of the token weighting in Eq.~(\ref{eq:token_weights}), we vary $w_{\text{silence}}^{\text{repeated}} \in \{0.2,\, 0.5,\, 0.8\}$ while keeping $w_{\text{silence}}^{\text{first}} = 1$ and $w_{\text{response}}$ fixed, and evaluate on the QA and Narration tasks. As shown in Figure~\ref{fig:ablation}~(a), the overall average score increases as $w_{\text{silence}}^{\text{repeated}}$ grows, suggesting that a larger $w_{\text{silence}}^{\text{repeated}}$ reduces mistimed responses. Empirically, however, $w_{\text{silence}}^{\text{repeated}}$ cannot be pushed too high (e.g., to $1$), as this causes the model to overcommit to silence on tasks that require frequent responses (e.g., narration). We therefore strike a balance and adopt $w_{\text{silence}}^{\text{repeated}} = 0.8$ as the default, which enables the model to learn \emph{when} to respond and prevents an overly aggressive silence bias.

\begin{figure}
  \centering
  \includegraphics[width=0.7\textwidth]{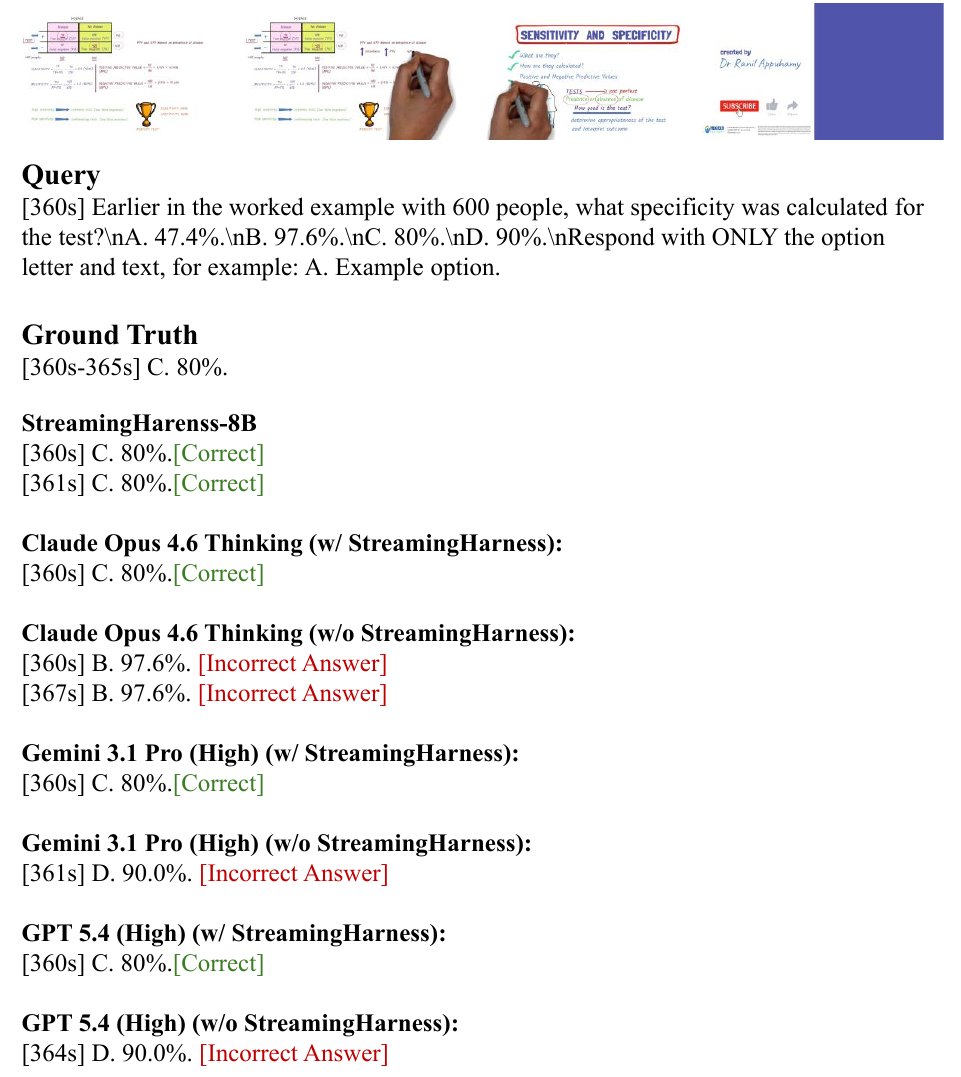} 
  \caption{Backward Task Example. }
  \label{fig:case4}
\end{figure}

To assess the effectiveness of our plug-and-play system, StreamingHarness, we compare models equipped with it (\textit{w/ StreamingHarness}) against their vanilla counterparts (\textit{w/o StreamingHarness}), where the vanilla setting retains only the most recent $T_s$ seconds of visual context, without mid-term or long-term memory. Figure~\ref{fig:ablation}~(b) reports the accuracy (TP) of four strong closed-source models on the \textit{Backward} task, which requires recalling distant evidence and thus relies heavily on long-horizon memory. The results show that equipping models with StreamingHarness yields substantial and consistent improvements across all four models, demonstrating its effectiveness as a general-purpose enhancement for long-horizon streaming understanding (see Figure~\ref{fig:case4}).
\section{Conclusion}

In this paper, we introduce a unified stack for streaming video understanding \textit{in the wild}, consisting of three components: \textbf{Streaming-Train-248K}, a large-scale dataset with per-second frame–text alignment that adapts off-the-shelf VLMs into streaming-native models;
\textbf{StreamingHarness}, a plug-and-play framework that equips any VLM with \textit{proactive interaction}, \textit{12-hour memory}, and \textit{sub-second latency}; 
and \textbf{Streaming-Eval}, a benchmark that jointly assesses answer correctness and response timeliness under realistic streaming scenarios. 
Extensive experiments demonstrate that our stack outperforms strong closed-source VLMs while maintaining stable sub-second latency across hours of continuous inference. 
Together, these contributions demonstrate the potential of deploying VLMs in real-world streaming applications, and advance the community's transition from offline video understanding toward deployable streaming intelligence.

\clearpage
\bibliographystyle{plainnat}
\bibliography{cite}

\newpage
\beginappendix

\section{Limitations}
\label{app:limitations}
Our current framework operates in a vision-language model setting and does not incorporate the audio modality. Integrating an omni-modal large model with speech understanding capabilities could further enhance narration quality, particularly for scenarios where audio cues (e.g., crowd reactions, referee whistles) carry critical information.

Second, to balance memory length and inference efficiency across the majority of real-world scenarios, we conduct our experiments at 1 FPS. However, certain domains demand higher temporal resolution to capture rapid visual changes, such as subtle facial expressions, fast-paced athletic movements, or brief but decisive moments in competitive sports. Our framework is not specifically optimized for these cases. In principle, one could reduce the memory length and increase the frame rate to accommodate such scenarios, but we leave a systematic exploration of this trade-off to future work.

\section{More Details of StreamingHarness}

\subsection{Prompt Template}
\label{appendix:prompt}
\begin{tcolorbox}[
  colback=black!5!white,
  colframe=black!75!black,
  title=System Prompt,
  fonttitle=\bfseries,
  breakable,
  enhanced
]
\small
You are a real-time streaming video agent. Each user turn contains the current timestamped frame, and earlier turns in the same chunk show recent visual context.

\medskip
\textbf{Output exactly one of:}
\begin{itemize}[leftmargin=*,itemsep=2pt,topsep=2pt]
  \item \texttt{</silence>}
  \item \texttt{</response>} one concise answer or commentary sentence
\end{itemize}

\medskip
\textbf{Rules:}
\begin{itemize}[leftmargin=*,itemsep=2pt,topsep=2pt]
  \item Your output must begin with \texttt{</silence>} or \texttt{</response>}.
  \item Use \texttt{</silence>} when no user query is active or when there is no useful new response for the active query.
  \item When a user query is active, answer that query using the current visual evidence, recent chunk context, and provided video memory.
  \item Do not invent names, scores, labels, or facts that are not visible or provided in the prompt.
\end{itemize}
\end{tcolorbox}

\newpage
\section{Evaluation}
\subsection{ Narration }
\label{appendix:narration}

\begin{tcolorbox}[
  colback=black!5!white,
  colframe=black!75!black,
  title=LLM-as-Judge Prompt,
  fonttitle=\bfseries,
  breakable,
  enhanced
]
\small
You are a strict judge of live-video narration / commentary.
You will see a human reference commentary (time-stamped) and two candidate commentaries A and B produced on the same video. Decide which candidate as a whole better matches the reference.

\medskip
Criteria, equally important:

\begin{enumerate}[leftmargin=*,itemsep=2pt,topsep=2pt]
  \item \textbf{Narration style (pacing + tone)} --- real commentators stay SILENT most of the time and speak only at key moments in natural spoken language. Candidates that narrate every second or read like written descriptions are more like captioning than narration.
  \item \textbf{Consistency \& accuracy} --- every detail stays synchronized with the reference and never contradicts it; no hallucinated facts, covers the key moments.
\end{enumerate}

\medskip
If a candidate is empty, the other wins unless both are empty (tie). If both candidates are roughly equally good or equally bad, output \texttt{tie}.

\medskip
Return STRICT JSON on one line:\\
\texttt{\{"winner": "A"|"B"|"tie", "reason": "<=500 words"\}}

\medskip
\textbf{[REFERENCE COMMENTARY]}\\
\texttt{\{gt\_transcript\}}

\medskip
\textbf{[CANDIDATE A]}\\
\texttt{\{a\_pred\}}

\medskip
\textbf{[CANDIDATE B]}\\
\texttt{\{b\_pred\}}
\end{tcolorbox}

\section{Case Study}

\end{document}